\documentclass[runningheads]{llncs}
\usepackage{cancel}
\usepackage{ulem}
\usepackage{color}
\usepackage{graphicx}
\usepackage{marvosym}
\usepackage{microtype}
\usepackage{subfigure}
\usepackage{booktabs} % for professional tables
\usepackage{floatrow}
\usepackage{times}
\usepackage{soul}
\usepackage{url}
\usepackage{float}
\usepackage[ruled,linesnumbered]{algorithm2e}
\usepackage{amsmath}
\usepackage[section]{placeins}
\usepackage{appendix}
\usepackage{amsmath}    %
\usepackage{booktabs}
\usepackage{algorithmic}
\usepackage{amssymb} %
\usepackage[utf8]{inputenc}
\usepackage{newunicodechar}  %
\usepackage{multirow}
\usepackage{ulem}
\newunicodechar{ℝ}{\mathbb{R}}
\usepackage{graphicx}
\usepackage{marvosym}
\usepackage{microtype}
\usepackage{graphicx}
\usepackage{subfigure}
\usepackage{booktabs} % for professional tables
\usepackage{floatrow}
\usepackage{times}
\usepackage{soul}
\usepackage{url}
\usepackage[hidelinks]{hyperref}
\usepackage[utf8]{inputenc}
\usepackage[small]{caption}
\usepackage{graphicx}
\usepackage{amsmath}
\usepackage{booktabs}
\usepackage{algorithmic}
\newfloatcommand{capbtabbox}{table}[][\FBwidth]
\begin{document}
\title{Multi-View Spatial-Temporal Network for Continuous  Sign Language Recognition}
\author{Ronghui Li \and
Lu Meng\textsuperscript{(\Letter)}
}
\authorrunning{R. Li and L. Meng}
\institute{Northeastern University, Shenyang 110000, China \\
\email{\{ronghuili16,menglu1982\}@gmail.com}}
\maketitle              % typeset the header of the contribution
\begin{abstract}
Sign language is a beautiful visual language and is also the primary language used by speaking and hearing-impaired people. However, sign language has many complex expressions, which are difficult for the public to understand and master. Sign language recognition algorithms will significantly facilitate communication between hearing-impaired people and normal people.
Traditional continuous sign language recognition often uses a sequence learning method based on Convolutional Neural Network (CNN) and Long Short-Term Memory Network (LSTM). These methods can only learn spatial and temporal features separately, which cannot learn the complex spatial-temporal features of sign language. LSTM is also difficult to learn long-term dependencies. To alleviate these problems, this paper proposes a multi-view spatial-temporal continuous sign language recognition network. The network consists of three parts. The first part is a Multi-View Spatial-Temporal Feature Extractor Network (MSTN), which can directly extract the spatial-temporal features of RGB and skeleton data; the second is a sign language encoder network based on transformer, which can learn long-term dependencies; the third is a Connectionist Temporal Classification (CTC) decoder network, which is used to predict the whole meaning of the continuous sign language. Our algorithm is tested on two public sign language datasets SLR-100 and PHOENIX-Weather 2014T (RWTH). As a result, our method achieves excellent performance on both datasets. The word error rate on the SLR-100 dataset 1.9$\%$, and the word error rate on the RWTHPHOENIX-Weather dataset is 22.8$\%$.

\end{abstract}
\section{Introduction}
Sign language is a conventional visual language, so it is different in every country and region. Sign language mainly uses hand information, movement trajectories, and gesture shapes, supplemented by facial expressions to convey information. It is the most important way for people with hearing impairment and speech impairment to communicate and learn knowledge in daily life. However, sign language is complex and difficult to learn. Therefore, video sign language recognition is becoming more and more important, which can promote better communication between the hearing impaired and the normal hearing people.

There is a big difference between continuous sign language recognition and isolated sign language recognition. Isolated sign language recognition is a classification task ~\cite{guo2017online,huang2018attention,yin2016iterative,zhang2016chinese}. These tasks are similar to action recognition. But isolated sign language recognition is far more complicated than action recognition. This is because sign language actions involve complicated finger interactions and excessive movement of multiple actions, and some sign language vocabulary are very similar. Action recognition only needs to recognize simple actions such as sitting down, bending, etc. Continuous sign language recognition is more complicated than isolated sign language recognition. This is because we need to recognize a series of consecutive sign words, and there is no obvious separation between each word. That is to say, the annotation information of continuous sign language recognition lacks the separation between words, so it belongs to the weakly supervised sequence to sequence task. 

In order to promote the development of continuous sign language recognition technology, some researchers have produced large-scale sign language datasets ~\cite{camgoz2018neural,huang2018video,koller2015continuous}. Methods based on deep learning have gradually become the mainstream method in recent years. These methods have achieved excellent performance in continuous sign language recognition tasks. Cui et al. ~\cite{cui2017recurrent} proposed recursive convolutional neural network segmentation optimization to recognize continuous sign language. Another study ~\cite{huang2018video} also showed the superiority of deep learning over manual feature-based methods in hierarchical attention in the latent space. 

We designed our continuous sign language recognition network into three parts: feature extraction, encoding network, and decoding network. With the development of deep neural network, we could extract depth images, optical flow and skeleton data from RGB frames. Single-modal data has some natural drawbacks: RGB images are easily affected by light and angle; skeleton data lacks facial details, etc. Effective multi-modal data fusion methods can highlight the merits of different modal data, and utilize the specificity and complementarity of multiple modal data to obtain robust sign language representations. At present, only a few continuous sign language recognition algorithms used multi-modal data fusion stratesgies~\cite{STMCTRANS}. In this work, we used the method of fusion skeleton and RGB data, which can not only take advantages of the good robustness of skeleton data, but also take advantages of the rich information of RGB data.

We first used a sliding window to split the sign language data into clips, which can reduce the length of the input sequence and avoid the network from learning too long-term dependencies. The clip data contains spatial-temporal features, which are more conducive to continuous sign language recognition. We used the transformer-based Vision Transformer Network (ViT) ~\cite{VIT32}  to extract RGB clips' spatial-temporal features and designed an Attention-enhanced Multi-scale 3D Graph Convolutional Network (AM3D-GCN) as the feature extractor of the skeleton clips. This is because the skeleton data could be regarded as natural graph data. Using the GCN network can learn the coordinate information of each joints and the relationship information between the joints. Then, the use of 3D-GCN can directly learn spatial-temporal features. The use of multi-scale architecture can allow the network to learn the dependencies between long-distance joints better. The use of spatial-temporal attention mechanism can further improve the robustness of the network and reduce the negative effects of imprecise points on recognition results.

In the Encoder network part, the previous methods mainly used RNN and its variant  network~\cite{pu2019}. However, sign language sequences often contain hundreds of frames, and it is difficult for RNNs to learn such long-term dependence. We designed an encoder network based on Transformer~\cite{transformer}, which has the advantages of high efficiency, suitable for long-sequence modelling, and self-attention. In the Decoder network, we designed a CTC decoder network. Finally, the sign language words sequence can be identified from the continuous sign language skeleton data and RGB data.

In general, our contributions are mainly reflected in the following aspects:

1. We used the feature fusion strategy of RGB and skeleton to further improve recognition performance, which can take advantage of the robustness of the skeleton data and the valuable information of RGB data. A sliding window is designed to reduce the input sequence length, which is convenient for the subsequent network to extract spatial-temporal features.

2. In the feature extractor part of the skeleton data, we designed an AM3D-GCN network. Its 3D architecture can directly extract spatial-temporal features; the design of multi-scale parallel GCN enables it to learn features at different levels; the spatial-temporal attention network can further improve its robustness.

3. We designed a continuous  sign language encoder
network based on Transformer and designed a CTC decoder network. Furthermore, we conducted sufficient experiments on two public sign language datasets, and the recognition results reached the state of art.

\section{Related work}

\subsection{Skeleton based Methods in Isolated Sign Language Recognition}
Many researchers used skeleton data for isolated sign language recognition. For example, Devineau et al.~\cite{devineau2018deep} used skeletal data to generate fake images and then used Convolution Neural Network (CNN) to process them. Konstantinidis et al.~\cite{konstantinidis2018sign} regarded the skeleton data as sequence data and used the Recurrent neural network (RNN) methods to process it. However, the recognition results of the above algorithm were unsatisfactory. This is because when the skeleton data is processed as images or sequence data, the association relationship between the joints is lost.

Skeleton data is natural graph data, so it is very suitable for feature extraction with GCN.
SAM-SLR~\cite{jiang2021skeleton} used the fusion of skeleton data, RGB data, optical flow data, and depth data for sign language vocabulary recognition and they designed SL-GCN to extract the skeleton data features.
SLR-Net~\cite{Rong2021attention} is a GCN algorithm based only on skeleton data. It has achieved the best recognition effect on two public datasets CSL-500~\cite{zhang2016chinese} and DEVISIGN-L~\cite{devisign}.

\subsection{Continuous Sign Language Recognition}
In recent years, 
continuous sign language recognition methods based on deep neural networks~\cite{cihan2017subunets,cui2017recurrent} mostly used the CNN-LSTM network architecture. These methods used CNN to extract spatial features, then used LSTM to learn temporal features. There were also some methods that combine traditional hidden Markov models with deep neural networks. For example, Koller et al.~\cite{koller2017re} embedded HMM into the CNN-LSTM network.

\section{Method}

\begin{figure}
\includegraphics[width=\textwidth]{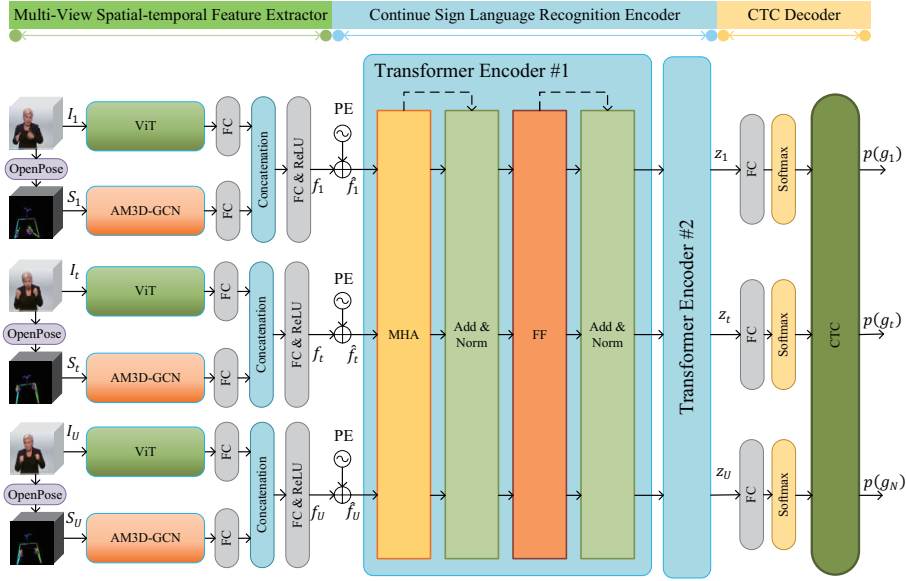}
\caption{Overview of Multi-view Spatial-temporal Continuous Sigan Language Recognition Network.PE means Position Encoding. MHA measn Multi Head Attention Network. FC means Fully Connection Layer.} \label{fig1}
\end{figure}

As shown in Fig.1. Given a sign video $X=(x_1,x_2,..,x_T)$ with T frames, our goal is to generate a sign glosses sequence $G=(g_1,g_2,..,g_N)(T>>N)$. This means learning conditional probability $p(G|X)$ in mathematical terms. In addition, our source sequence is video. We first used a sliding window to segment the video, which greatly reduces the input sequence length. We then designed a Multi-View Spatial-Temporal Feature Extractor to extract spatial-temporal features $f_{1:U}$ of RGB and skeleton clips. In addition, we used a transformer-based architecture to get the representations $z_{1:U}$ of the sequence at different time steps. Finally, We trained our method in an end-to-end manner with CTC loss function and get the output glosses sequences.

\begin{figure}[ht]
\centering
\includegraphics[scale=0.8]{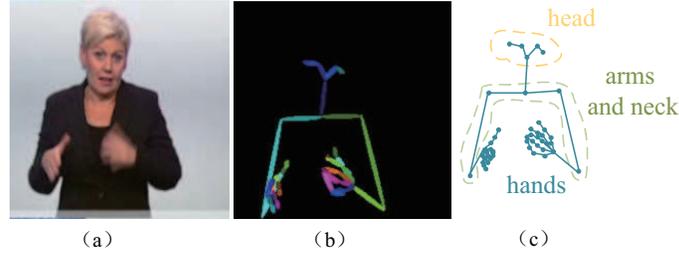}
\caption{Example of multi-modal data. (a) RGB data. (b) Skeleton data displayed as a colorful image. (c) Skeleton data displayed as lines and points.}
\label{fig:label}
\end{figure}

\subsection{MSTFE: Multi-View Spatial-Temporal Feature Extractor}
First, the skeleton data was extracted from the RGB video using the OpenPose~\cite{openposebody} algorithm. As shown in Fig.2, a total of 52 joints were used, including 42 joints for the hands, 5 joints for the face, and 5 joints for arms and neck. Then the skeleton data was preprocessed by geometric translation and zooming to reduce the differences caused by the height and body of different people. We used skeleton data and RGB data to perform sign language recognition together.

Given a sign video $X$, we first used a sliding window to generate a set of ordered video clips. The sliding window length has been set to $m$, which means each clip has $m$ frames. 
RGB clip is represented as $I=(I_1,\cdots,I_U)=\left\{I_t\right\}_{t=1}^U$ with U clips, and the skeleton clip set is $S=(S_1,\cdots,S_U)=\left\{S_t\right\}_{t=1}^U$. The input of MSTFE is RGB clips and skeleton clips. The spatial-temporal features of RGB clips were extracted by ViT~\cite{VIT32}, and the spatial-temporal features of skeleton clips were extracted by AM3D-GCN. The features of these different modalities are respectively processed by the FC layer, so that the output dimensions are all 512 dimensions. Then the two features are concatenated to obtain 1024-dimensional features, and then fully connection and ReLU operations are performed on them to obtain the multi-view spatial-temporal representations of $f_{1:U}$, where $f_t\epsilon\mathbb{R}^d,\ d=1024$.

\begin{figure}[ht]
\centering
\includegraphics[scale=0.6]{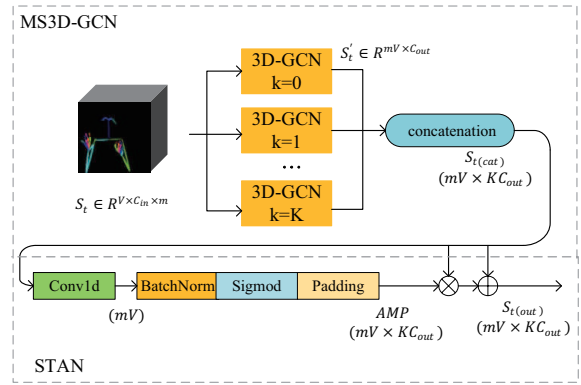}
\caption{The structure of AM3D-GCN. The network consists of a Multi-scale 3D GCN (MS3D-GCN) and a Spatial-Temporal Attention Network (STAN)."$\bigoplus$" means add operation. $\bigotimes$" means matrix dot-product operation.}
\label{fig:label}
\end{figure}

\subsection{AM3D-GCN: Attention-enhanced Multi-scale 3D GCN}
As shown in Fig.3, AM3D-GCN is divided into Multi-Scale 3D GCN (MS3D-GCN) and Spatial-Temporal Attention Network (STAN). The dimension of the input data skeleton Clip is $S_t\in R^{V\times C_{in}\times m}$, where $V$ is the number of joints, $C_{in}=2$ represents the two-dimensional point coordinates, and $m$ is the number of frames.

\subsubsection{Traditional 2D-GCN} Traditional 2D-GCN could extract the spatial features of each skeleton frames. The graph data of 2D-GCN can be built as $G=\left(V,\varepsilon\right)$, where $V=\left(V_1,V_2,\cdots V_v\right)$ is the set of joints, $\varepsilon$ is the set of edges which means the connection between the joint. $\varepsilon$ is the set of edges, which can be expressed by the adjacency matrix of $\widetilde{A}\in\mathbb{R}^{V\times V}$.

\begin{equation}
  \begin{split}
  &A_{(k)i,j} = \left\{
            \begin{array}{lr}
             1,if\ d\left(V_i,\ V_j\right)=1 &  \\
             0, otherwise
             \end{array}
\right.,    \\
  &\widetilde{A}=A+I.
  \end{split}
  \end{equation}

Where $d\left(V_i,V_j\right)$ is the shortest distance on the path of human bones between the two joints $V_i$ and $V_j$.

\subsubsection{3D-GCN} Based on 2D-GCN, we designed a 3D-GCN that can directly extract spatial-temporal features. The 3D graph $G_{\left(m\right)}=\left(v_{\left(m\right)},\varepsilon_{\left(m\right)}\right)$, where $V_{(m)}=\left(V_1,V_2,\cdots V_{mv}\right)$ is the joints set of m frames. The edge set $\varepsilon_{\left(m\right)}$ is defined by tiling $\widetilde{A}$ into a block adjacency matrix${\widetilde{A}}_{\left(m\right)}$.

\begin{equation}
	{\widetilde{A}}_{\left(m\right)}=\left[\begin{matrix}{\widetilde{A}}&\cdots&{\widetilde{A}}\\\vdots&\ddots&\vdots\\{\widetilde{A}}&\cdots&{\widetilde{A}}\\\end{matrix}\right]\epsilon R^{m V\times m V}
\end{equation}
Equation (2) means that the j-th joint of one frame will be connected to the one-hop neighbor joints and all the j-th joint in m frames. We can get 3D-GCN as Equation (3). The dimension of $S_t$ is $(m; Cin; V )$. In order to use equation 2, we resized $S_t$ as $(Cin; mV )$.
\begin{equation}
	S_{t}^\prime=\sigma({\widetilde{D}}_{\left(m\right)}^{-\frac{1}{2}}{\widetilde{A}}_{\left(m\right)}{\widetilde{D}}_{\left(m \right)}^{-\frac{1}{2}}S_{t}W_k)
\end{equation}

Where ${\widetilde{A}}_{\left(m\right)}$ is the adjacency matrix, defined by (4). ${\widetilde{D}}_{\left(m\right)}$ is diagonal degree matrix of ${\widetilde{A}}_{\left(m\right)}$. $W_k$ means a trainable weight matrix of network. $\sigma()$ means ReLU activation function.

\subsubsection{Multi scale 3D-GCN} The previous GCN algorithms were inefficient in learning the dependencies between long-distance joints. However, the relationship between long-distance joints, such as left and right fingers, is often more important for sign language recognition. Therefore, we designed a multi-scale 3D-GCN to allow the network to learn the dependencies between distant nodes. As in Equation (4), we constructed $K$ different adjacency matrices by adjusting the scale k to learn features at different distance levels:

\begin{equation}
  \begin{split}
  &A_{(k)i,j} = \left\{
            \begin{array}{lr}
             1,if\ d\left(V_i,\ V_j\right)=k &  \\
             0, otherwise
             \end{array}
\right.    \\
  &\widetilde{A}_{(k)}=A_{(k)}+I   \\
  &{\widetilde{A}}_{\left(k,m\right)}=\left[\begin{matrix}{\widetilde{A}}_{\left(k\right)}&\cdots&{\widetilde{A}}_{\left(k\right)}\\\vdots&\ddots&\vdots\\{\widetilde{A}}_{\left(k\right)}&\cdots&{\widetilde{A}}_{\left(k\right)}\\\end{matrix}\right]\epsilon R^{m V\times m V}
  \end{split}
  \end{equation}
Multi-scale 3D-GCN can be expressed by the following Equation:
\begin{equation}
	S_{t(cat)}=cat\left[{\sigma\left({\widetilde{D}}_{\left(k,m\right)}^{-\frac{1}{2}}{\widetilde{A}}_{\left(k,m\right)}{\widetilde{D}}_{\left(k,m\right)}^{-\frac{1}{2}}S_tW_k\right)}_{k=1}^K\right]
\end{equation}

Where cat[] means concatenation operator.

\subsubsection{STAN: Spatial-Temporal Attention Network} $S_{t(cat)}$ were sent to STAN, which can further enhance the robustness of the network and reduce the negative effects of imprecise joints. STAN is a Spatial-Temporal attention network, after the operation of Conv1d, the dimension of $S_{t(cat)}$ became $\left(mV\right)$. Padding means copy the feature at the channel dimension to get the attention map (AMP) and $S_{t\left(out\right)}$ could be calculated as Equation (6).
\begin{equation}
	S_{t\left(out\right)}=S_{t\left(cat\right)}+S_{t\left(cat\right)}\bullet AMP
\end{equation}

Where "$\bullet$" means matrix dot-product operation.

\subsection{CSLEN: Continuous Sign Language Encoder Network} 
The Continuous Sign Language Encoder Network was designed based on the transformer encoding network. MSTFE has learned the feature $f_{1:U}$, which is obtained after position encoding ${\hat{f}}_{1:U}$. Among them, Position Encoding can generate different vectors according to time step, which is a predefined sine function. The inputs of CSLEN were modelled by a multi head attention layer. Outputs of the multi head attention layer were then passed through a non-linear point wise feed forward layer. There are also residual connections and normalization operations to help training. The output of Transformer encoding network is $Z^U=\left(z_1,\cdots,z_U\right)$, which is generated by CSLEN at time step $t$, given the multi-view spatial-temporal representations of all the video clips, ${\hat{f}}_{1:U}$.

\subsection{CTC Decoder Network} 
CTC can directly optimize the network for segmented sequence data without the need for segmentation and labeling between words, so it is suitable for processing weakly supervised continuous sign language recognition task. The sequence of continuous sign language and glosses sequence is the same, but sign language is much longer than glosses sequence, so one gloss often corresponds to multiple sign clips.
Given spatial-temporal representations, $z_{1:U}$ , we first used a FC layer to process, connected to the softmax activation function, then we used CTC to compute $p\left(G|I\right)$ by marginalizing over all possible $I$ to $G$ alignments as:

\begin{equation}
	p\left(G|I\right)=\sum_{\pi\in B} p\left(\pi|I\right)
\end{equation}
where $\pi$ is a path and $B$ are the set of all viable paths that correspond to G. We then use the $p\left(G^\ast|I\right)$ to calculate the CTC loss as:
\begin{equation}
	\mathcal{L}=1-p\left(G^\ast|I\right)
\end{equation}

where $G^\ast$ is the ground truth glosses sequence.

\section{Experiments} 
In this section, we conducted a series of ablation experiments to verify the effectiveness of our multi-modal data fusion strategy and AM3D-GCN. We then compared other SOTA methods, and the results showed that our method performed best on the SLR-100 dataset.
\subsection{Evaluation Metric} 
In continuous SLR, word error rate (WER) is the most widely-used metric to evaluate the performance. WER is essentially an edit distance. In other words, WER indicates the least operations of substitution, insertion, and deletion to transform the predict sentence into the reference sequence:
\begin{equation}
	WER=\frac{\#substitution+\#insertion+\#deletion}{length\ of\ reference}
\end{equation}

\subsection{Datasets} 
PHOENIX-Weather 2014T (RWTH) dataset was recorded by nine sign language speakers, with a vocabulary of 1066. It contains 7,096 training pairs, 519 development pairs, and 642 test pairs. The image resolution is 210x260. The continuous SLR dataset  (SLR-100) was executed by 50 signers. There are a total of 5,000 video-sentence pairs. The vocabulary is 178. The image resolution is 1280x720.

\subsection{Implementation detail} 
All components of our network were built by PyTorch~\cite{pytorch}. We conducted a sliding window on raw videos to generate clips. The window size was set to be 8 with a stride of 4, which means there was 50$\%$ overlap between adjacent clips. We also utilized 0.1 dropout rate on MSTFE and transformer layers to mitigate over-fitting. We used the Adam~\cite{kingma2014adam} optimizer, the batch size was set to 32, the learning rate and weight decay were set to $10^{-3}$.

\subsection{Ablation experiments} 
\subsubsection{Ablation Experiments of different model data}   We used ablation experiments to verifie the effectiveness of data in different modes. As shown in Table 1, thanks to the powerful feature extraction capabilities of AM3D-GCN and the refinement of skeleton data, using skeleton data on the SLR-100 dataset is better than that of RGB data. The use of multi-modal fusion methods can also effectively reduce WER.

\begin{table}
\centering
\caption{The ablation experiments results of the input data with different models. Tested on SLR-100 dataset.}\label{tab1}
\begin{tabular}{|l|l|l|}
\hline
RGB clips &  skeleton clips & WER\\
\hline
$\surd$ &   & 3.2\\
 \hline
 & $\surd$  & 2.1\\
  \hline
$\surd$ & $\surd$ & 1.9\\
\hline
\end{tabular}
\end{table}

\subsubsection{Ablation Experiments of AM3D-GCN}  In this section, we will further verify the effectiveness of the 3D-GCN, Multi-Scale and STAN in the AM3D-GCN. In this part of the experiment, only the skeleton data was input into the network.  As shown in Table 2. The first row means using 2D-GCN to extract the spatial representations of each skeleton frame without using sliding windows to segment the clips. It can be seen that the WER is the highest, reaching 3.6$\%$. Analyzing this experimental data, we know that in the continuous sign language recognition task, extracting the spatial-temporal features of clips is better than extracting the spatial features of single-frame data. Then we tested the performance improvement of Multi-Scale and STAN, of which Multi-Scale performed even better. It can be seen that a well-designed feature extraction network can significantly improve the performance of the continuous sign language recognition network.
\begin{table}
\centering
\caption{The ablation experiments results of the architecture of AM3D-GCN. Tested on SLR-100 dataset.}\label{tab1}
\begin{tabular}{|l|l|l|l|l|l|}
\hline
Sliding Window &  2D-GCN & 3D-GCN & Multi-Scale & STAN & WER\\
\hline
 &  $\surd$ &  &   &  & 3.6\\
 \hline
$\surd$ &  & $\surd$ &   &  & 2.8\\
\hline
$\surd$ &  & $\surd$ &  $\surd$ &  & 2.4\\
\hline
$\surd$ &  & $\surd$ &   & $\surd$ & 2.7\\
\hline
$\surd$ &  & $\surd$ &  $\surd$ & $\surd$ & 2.1\\
\hline
\end{tabular}
\end{table}

\subsection{Compared with other SOTA methods}
In this section, we compared our algorithm with other advanced continuous sign language recognition algorithms. Our method ranked first on the SLR-100 dataset and second on the RWTH dataset. This may be due to the low image quality of the RWTH dataset, which makes it is difficult to extract precise joints from it. STMC is also a method based on multi-modal data and has also shown outstanding perfor-mance. It can be seen from Table 4 that SLRT, STMC and our method are all based on the Transformer architecture, and their performance is much better than previous models. In summary, multi-modal data and Transformer architecture can effec-tively reduce the WER of continuous sign language recognition.
\begin{table*}
\begin{floatrow}
\capbtabbox{
 \begin{tabular}{|l|l|}
 \hline
  Method & WER \\
 \hline
%DTW-HMM[47]  &  28.4\\
%\hline
%LSTM [38]  &  26.4\\
%\hline
%S2VT [37]  &  25.5\\
%\hline
%LSTM-A [45]  &  24.3\\
%\hline
%LSTM-E [28]  &  23.2\\
%\hline
HAN~\cite{yang2016continuous}   &  20.7\\
\hline
LS-HAN~\cite{huang2018video}   &  17.3\\
\hline
SubUNet~\cite{cihan2017subunets}   &  11.0\\
\hline
HLSTM~\cite{guo2018hierarchical}   &  7.6\\
\hline
HLSTM-attn~\cite{guo2018hierarchical}   &  7.1\\
\hline
Align-iOpt~\cite{pu2019}   &  6.1\\
\hline
SF-Net~\cite{yang2019sf}    &  3.8\\
\hline
FCN~\cite{FCN2020}   &  3.0\\
\hline
STMC~\cite{STMC_CSL}   &  2.1\\
\hline
Ours   &  1.9\\
 \hline
 \end{tabular}
}
{
 \caption{Results comparison on SLR-100.}
 \label{tab:tb1}
}
\capbtabbox{
 \begin{tabular}{|l|l|}
 \hline
  Method & WER \\
%\hline
%Deep Hand~\cite{koller2016deephand}  &  45.1\\
\hline
Deep Sign~\cite{koller2016deepsign}  &  38.8\\
\hline
SubUNet~\cite{cihan2017subunets}  &  40.7\\
\hline
Staged-Opt~\cite{cui2017recurrent}  &  38.7\\
\hline
LS-HAN~\cite{huang2018video}   &  38.3\\
\hline
Align-iOpt~\cite{pu2019}   &  36.7\\
\hline
SF-Net~\cite{yang2019sf}    &  38.1\\
\hline
SLRT~\cite{slt}   &  24.6\\
\hline
FCN~\cite{FCN2020}   &  23.9\\
\hline
STMC~\cite{STMCTRANS}   &  21.1\\
\hline
Ours   &  22.8\\
 \hline
 \end{tabular}
}{
 \caption{Results comparison on RWTH.}
 \label{tab:tb2}
}
\end{floatrow}
\end{table*}

\subsection{Recognition examples}
We gave two recognition examples as shown in Fig.4. The original datasets were in Chinese and German respectively. In order to facilitate understanding, we manually translated the predicted results and ground truth into English.

\begin{figure}[ht]
\centering
\includegraphics[scale=0.6]{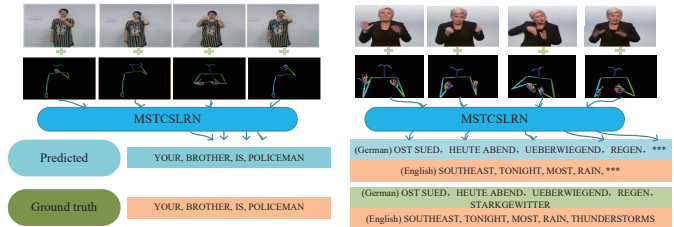}
\caption{Example results of our method. MSTCSLRN:Multi-View Spatial-Temporal Continuous Sign Language Recognition Network}
\label{fig:label}
\end{figure}

\section{Conclusion}

In this paper, we designed a novel and effective continuous sign language recognition network, which can be divided into three parts: Multi-View feature Extractor, Continuous Sign Language Encoder Network, and CTC decoder Network. In the past, continuous sign language recognition networks paid more attention to the sequence model and ignored the importance of the feature extraction network. This paper designs a feature extraction network that can simultaneously learn the spatial-temporal representation of RGB and skeleton data. We designed AM3D-GCN to model skeleton clips. It contains 3D-GCN that can effectively learn the spatial-temporal representations, and the multi-scale mechanism can learn the relationship between different distance levels of joints. The spatial-temporal attention network is added to improve its robustness and reduce the negative effects of imprecise joints on the recognition results. Finally, we have designed sufficient experiments to verify the effectiveness of each part of our network. Our method ranked first on the SLR-100 dataset. 

%citations of references, we prefer the use of square brackets
%and consecutive numbers. Citations using labels or the author/year
%convention are also acceptable. The following bibliography provides
%a sample reference list with entries for journal
%articles~\cite{ref_article1}, an LNCS chapter~\cite{ref_lncs1}, a
%book~\cite{ref_book1}, proceedings without editors~\cite{ref_proc1},
%and a homepage~\cite{ref_url1}. Multiple citations are grouped
%\cite{ref_article1,ref_lncs1,ref_book1},
%\cite{ref_article1,ref_book1,ref_proc1,ref_url1}.

\bibliographystyle{splncs04}
\bibliography{mybibliography}

\begin{thebibliography}{10}
\providecommand{\url}[1]{\texttt{#1}}
\providecommand{\urlprefix}{URL }
\providecommand{\doi}[1]{https://doi.org/#1}

\bibitem{camgoz2018neural}
Camgoz, N.C., Hadfield, S., Koller, O., Ney, H., Bowden, R.: Neural sign
  language translation. In: Proceedings of the IEEE Conference on Computer
  Vision and Pattern Recognition. pp. 7784--7793 (2018)

\bibitem{slt}
Camgoz, N.C., Koller, O., Hadfield, S., Bowden, R.: Sign language transformers:
  Joint end-to-end sign language recognition and translation. In: Proceedings
  of the IEEE/CVF Conference on Computer Vision and Pattern Recognition. pp.
  10023--10033 (2020)

\bibitem{openposebody}
Cao, Z., Hidalgo, G., Simon, T., Wei, S.E., Sheikh, Y.: Openpose: realtime
  multi-person 2d pose estimation using part affinity fields. IEEE transactions
  on pattern analysis and machine intelligence  \textbf{43}(1),  172--186
  (2019)

\bibitem{FCN2020}
Cheng, K.L., Yang, Z., Chen, Q., Tai, Y.W.: Fully convolutional networks for
  continuous sign language recognition. In: European Conference on Computer
  Vision. pp. 697--714. Springer (2020)

\bibitem{cihan2017subunets}
Cihan~Camgoz, N., Hadfield, S., Koller, O., Bowden, R.: Subunets: End-to-end
  hand shape and continuous sign language recognition. In: Proceedings of the
  IEEE International Conference on Computer Vision. pp. 3056--3065 (2017)

\bibitem{cui2017recurrent}
Cui, R., Liu, H., Zhang, C.: Recurrent convolutional neural networks for
  continuous sign language recognition by staged optimization. In: Proceedings
  of the IEEE Conference on Computer Vision and Pattern Recognition. pp.
  7361--7369 (2017)

\bibitem{devineau2018deep}
Devineau, G., Moutarde, F., Xi, W., Yang, J.: Deep learning for hand gesture
  recognition on skeletal data. In: 2018 13th IEEE International Conference on
  Automatic Face \& Gesture Recognition (FG 2018). pp. 106--113. IEEE (2018)

\bibitem{VIT32}
Dosovitskiy, A., Beyer, L., Kolesnikov, A., Weissenborn, D., Zhai, X.,
  Unterthiner, T., Dehghani, M., Minderer, M., Heigold, G., Gelly, S., et~al.:
  An image is worth 16x16 words: Transformers for image recognition at scale.
  arXiv preprint arXiv:2010.11929  (2020)

\bibitem{guo2017online}
Guo, D., Zhou, W., Li, H., Wang, M.: Online early-late fusion based on adaptive
  hmm for sign language recognition. ACM Transactions on Multimedia Computing,
  Communications, and Applications (TOMM)  \textbf{14}(1),  1--18 (2017)

\bibitem{guo2018hierarchical}
Guo, D., Zhou, W., Li, H., Wang, M.: Hierarchical lstm for sign language
  translation. In: Proceedings of the AAAI Conference on Artificial
  Intelligence. vol.~32 (2018)

\bibitem{huang2018attention}
Huang, J., Zhou, W., Li, H., Li, W.: Attention-based 3d-cnns for
  large-vocabulary sign language recognition. IEEE Transactions on Circuits and
  Systems for Video Technology  \textbf{29}(9),  2822--2832 (2018)

\bibitem{huang2018video}
Huang, J., Zhou, W., Zhang, Q., Li, H., Li, W.: Video-based sign language
  recognition without temporal segmentation. In: Proceedings of the AAAI
  Conference on Artificial Intelligence. vol.~32 (2018)

\bibitem{jiang2021skeleton}
Jiang, S., Sun, B., Wang, L., Bai, Y., Li, K., Fu, Y.: Skeleton aware
  multi-modal sign language recognition. In: Proceedings of the IEEE/CVF
  Conference on Computer Vision and Pattern Recognition. pp. 3413--3423 (2021)

\bibitem{kingma2014adam}
Kingma, D.P., Ba, J.: Adam: A method for stochastic optimization. arXiv
  preprint arXiv:1412.6980  (2014)

\bibitem{koller2015continuous}
Koller, O., Forster, J., Ney, H.: Continuous sign language recognition: Towards
  large vocabulary statistical recognition systems handling multiple signers.
  Computer Vision and Image Understanding  \textbf{141},  108--125 (2015)

\bibitem{koller2016deepsign}
Koller, O., Zargaran, O., Ney, H., Bowden, R.: Deep sign: Hybrid cnn-hmm for
  continuous sign language recognition. In: Proceedings of the British Machine
  Vision Conference 2016 (2016)

\bibitem{koller2017re}
Koller, O., Zargaran, S., Ney, H.: Re-sign: Re-aligned end-to-end sequence
  modelling with deep recurrent cnn-hmms. In: Proceedings of the IEEE
  Conference on Computer Vision and Pattern Recognition. pp. 4297--4305 (2017)

\bibitem{konstantinidis2018sign}
Konstantinidis, D., Dimitropoulos, K., Daras, P.: Sign language recognition
  based on hand and body skeletal data. in 2018-3dtv-conference: The true
  vision-capture, transmission and display of 3d video (3dtv-con)(pp. 1-4)
  (2018)

\bibitem{Rong2021attention}
Meng, L., Li, R.: An attention-enhanced multi-scale and dual sign language
  recognition network based on a graph convolution network. Sensors
  \textbf{21}(4), ~1120 (2021)

\bibitem{pytorch}
Paszke, A., Gross, S., Chintala, S., Chanan, G., Yang, E., DeVito, Z., Lin, Z.,
  Desmaison, A., Antiga, L., Lerer, A.: Automatic differentiation in pytorch
  (2017)

\bibitem{pu2019}
Pu, J., Zhou, W., Li, H.: Iterative alignment network for continuous sign
  language recognition. In: Proceedings of the IEEE/CVF Conference on Computer
  Vision and Pattern Recognition. pp. 4165--4174 (2019)

\bibitem{transformer}
Vaswani, A., Shazeer, N., Parmar, N., Uszkoreit, J., Jones, L., Gomez, A.N.,
  Kaiser, L., Polosukhin, I.: Attention is all you need. arXiv preprint
  arXiv:1706.03762  (2017)

\bibitem{yang2016continuous}
Yang, W., Tao, J., Ye, Z.: Continuous sign language recognition using level
  building based on fast hidden markov model. Pattern Recognition Letters
  \textbf{78},  28--35 (2016)

\bibitem{yang2019sf}
Yang, Z., Shi, Z., Shen, X., Tai, Y.W.: Sf-net: Structured feature network for
  continuous sign language recognition. arXiv preprint arXiv:1908.01341  (2019)

\bibitem{yin2016iterative}
Yin, F., Chai, X., Chen, X.: Iterative reference driven metric learning for
  signer independent isolated sign language recognition. In: European
  Conference on Computer Vision. pp. 434--450. Springer (2016)

\bibitem{devisign}
Yin, F., Chai, X., Chen, X.: Iterative reference driven metric learning for
  signer independent isolated sign language recognition. In: European
  Conference on Computer Vision. pp. 434--450. Springer (2016)

\bibitem{STMCTRANS}
Yin, K., Read, J.: Better sign language translation with stmc-transformer. In:
  Proceedings of the 28th International Conference on Computational
  Linguistics. pp. 5975--5989 (2020)

\bibitem{zhang2016chinese}
Zhang, J., Zhou, W., Xie, C., Pu, J., Li, H.: Chinese sign language recognition
  with adaptive hmm. In: 2016 IEEE international conference on multimedia and
  expo (ICME). pp.~1--6. IEEE (2016)

\bibitem{STMC_CSL}
Zhou, H., Zhou, W., Zhou, Y., Li, H.: Spatial-temporal multi-cue network for
  continuous sign language recognition. In: Proceedings of the AAAI Conference
  on Artificial Intelligence. vol.~34, pp. 13009--13016 (2020)

\end{thebibliography}

\end{document}